\documentclass[10pt,twocolumn,letterpaper]{article}

\usepackage[pagenumbers]{cvpr} 

\usepackage[dvipsnames]{xcolor}

\definecolor{cvprblue}{rgb}{0.21,0.49,0.74}
\usepackage[pagebackref,breaklinks,colorlinks,citecolor=cvprblue]{hyperref}
\usepackage{tabularx}

\def\paperID{1590} 
\def\confName{CVPR}
\def\confYear{2024}

\title{Mitigating Hallucination in Visual Language Models with Visual Supervision}

\author{
  Zhiyang Chen $^{1,2}$ \quad Yousong Zhu $^{1}$ \quad Yufei Zhan $^{1,2}$ \quad Zhaowen Li $^{1,2}$ \quad Chaoyang Zhao $^{1}$ \\
  Jinqiao Wang $^{1,2,3,4}$ \quad Ming Tang $^{1}$\\
  Foundation Model Research Center, Institute of Automation, Chinese Academy of Sciences$^{1}$ \\ 
  School of Artificial Intelligence, University of Chinese Academy of Sciences$^{2}$\\
  Peng Cheng Laboratory$^{3}$ \quad  Wuhan AI Research$^{4}$ \\
{\tt\small \{zhiyang.chen,yousong.zhu,zhaowen.li,chaoyang.zhao,jqwang,tangm\}@nlpr.ia.ac.cn} \\
{\tt\small  zhanyufei2021@ia.ac.cn}
\vspace{-2em}
}

\begin{document}
\maketitle
\begin{abstract}
Large vision-language models (LVLMs) suffer from hallucination a lot, generating responses that apparently contradict to the image content occasionally.
The key problem lies in its weak ability to comprehend detailed content in a multi-modal context, which can be mainly attributed to two factors in training data and loss function. The vision instruction dataset primarily focuses on global description, and the auto-regressive loss function favors text modeling rather than image understanding.
In this paper, we bring more detailed vision annotations and more discriminative vision models to facilitate the training of LVLMs, so that they can generate more precise responses without encounter hallucination.
On one hand, we generate image-text pairs with detailed relationship annotations in panoptic scene graph dataset (PSG). These conversations pay more attention on detailed facts in the image, encouraging the model to answer questions based on multi-modal contexts.
On the other hand, we integrate SAM and mask prediction loss as auxiliary supervision, forcing the LVLMs to have the capacity to identify context-related objects, so that they can generate more accurate responses, mitigating hallucination.
Moreover, to provide a deeper evaluation on the hallucination in LVLMs, we propose a new benchmark, RAH-Bench. It divides vision hallucination into three different types that contradicts the image with wrong categories, attributes or relations, and introduces False Positive Rate as detailed sub-metric for each type.
In this benchmark, our approach demonstrates an +8.4\% enhancement compared to original LLaVA and achieves widespread performance improvements across other models.
\end{abstract}    
\vspace{-1em}
\section{Introduction}
\label{sec:intro}

Recently, large language models (LLMs) \cite{transformer,llama,vicuna,baichuan2,gpt4} have achieved significant successes in the field of Natural Language Processing (NLP). Due to the generative pretraining process \cite{gpt3} on a large amount of text corpus, these models have strong capacities on different language tasks. They are able to comprehend complex text inputs, and provide flexible responses when interacting with human. The succuess in text inspires the researchers to focus on understanding inputs of other modals, such as images. As a result, there appear Large Vision-Language Models (LVLMs) \cite{llava,blip2,instructblip}. LVLMs use a pretrained visual encoder \cite{clip,vit} to extract image features, and align them with a LLM via multimodal pretraining and instruction tuning. This training process makes LVLMs possible to conduct a complex conversation based on image content.
\begin{figure}
  \centering
  \includegraphics[width=\linewidth]{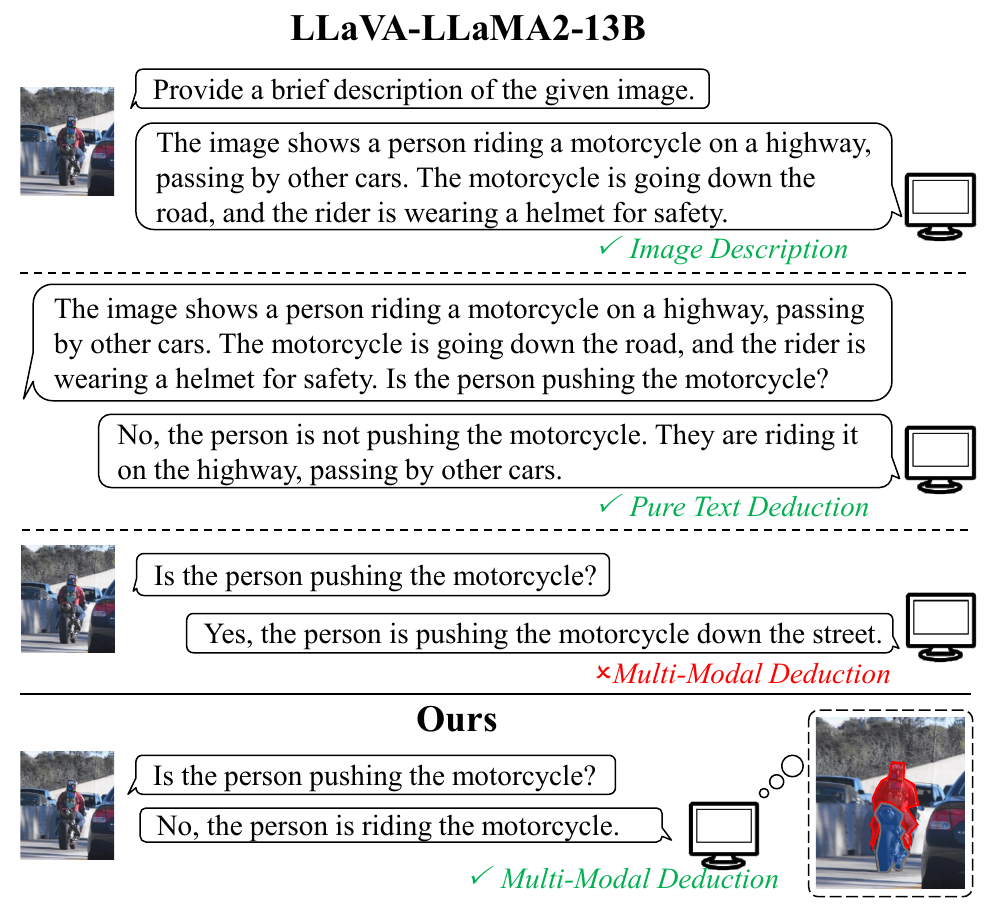}
  \caption{LVLMs are able to infer the correct answer from a pure-text context, yet struggle with a raw image. After training with visual supervision, LVLMs can generate more precise responses.}
  \label{fig:chat}
\end{figure}

Even though LVLMs inherit strong feature representation and generation capabilities from up-to-date LLMs \cite{llama2,vicuna} and image models \cite{clip}, they still generate unsatisfactory responses sometimes. One main problem is hallucination.
In the context of LLMs, hallucination refers to a phenomenon that the generated outputs are sometimes detached from, or even violate the provided inputs.
These responses are generated by merely following the learned patterns from training corpus, neglecting factual and accurate information in the provided text \cite{hallucination_survey}.
When taking the additional input image into consideration,
LVLMs might still generate outputs contrary to the image content.
Since the image itself provides sufficient content to conduct a wide range of visual tasks \cite{maskrcnn,fcn,sam,obj2seq},
the main problem is that the model lacks a comprehensive understanding of the multi-modal context.
As shown in Figure~\ref{fig:chat}, LVLMs have the capacity to describe an image or respond correctly to text questions, but generate hallucinations when answering directly according to the image.
We refer the inconsistency between the response and the given image as vision hallucination.
Hopefully, there appears some works trying to solve this problem. POPE \cite{pope} provides a new benchmark to evaluate object hallucination. LRV-Instruction \cite{lrv_instruction} introduces a more robust instruction dataset. Woodpecker \cite{woodpecker} performs inference one more time, so that it can correct hallucination based on more-detailed text prompts.
These methods mainly address hallucination from the aspect of text. On the contrary, we suggest improving the model's ability to comprehend the spatial structure and detailed relationships in the multi-modal context is also important.

However, we may wonder: with such abundant information in an image, why cannot a LVLM understand properly and generate the right response?
There are two causes that LVLMs miss the crucial features: lack of fine-grained alignment visual annotations and insufficient supervision to explicitly learn visual structures.
On one hand, existing visual instruction datasets \cite{llava,svit} generated by LLMs tend to focus on global descriptions, and contain mostly positive descriptions. These training data can hardly cover all potential statements about the input images, especially those misleading questions.
On the other hand, LVLMs are supervised with the next-token prediction loss, which is inherited from NLP to model the dependency among word tokens. It is unable to model the visual relationships and understand spatial regions in the image. Thus, it is hard to guarantee that LVLMs can answer a specific question according to the input image.

Therefore, in this paper, we construct a fine-grained vision instruction dataset based on Panoptic Scene Graph (PSG) \cite{psg}, called \emph{Relation-Associated Instruction (RAI-30k)}, which focuses on answering questions about detailed relations among instances. Other than standard dialogs, each instruction data in RAI-30k is also associated with one relation annotation in PSG, including mask annotations for related instances.
With these additional annotations, we further supervise LVLMs with mask prediction loss by a state-of-the-art expert vision model, guiding LVLMs to focus on highly-related image content.
More specifically, this is achieved by integrating SAM into the training of LVLMs. SAM receives the outputs from LVLMs, generates masks for instances associated with the instruction data. With the additional supervision from the mask prediction loss, LVLMs are encouraged to extract features that can better represents these crucial instances, thus generating more accurate responses and mitigating vision hallucination.
In addition, the proposed method only operates the training pipeline. The LVLMs still follow their original manner in inference, but with more precise outputs generated.
Moreover, in order to conduct a deeper evaluation on vision hallucination, we propose a new hallucination benchmark, called \emph{Relataion-Associated Hallucination Benchmark (RAH-Bench)}. It contains 3,000 interrogative sentences with corresponding images, and asks the model to judge if these detailed descriptions are consistent with image contents. To provide a detailed analysis on what mistakes LVLMs are more likely to make, all negative queries are divided into three types based on how they contradict the image: category hallucination, attribute hallucination and relation hallucination. For each type, we design a detailed sub-metric to reveal how vulnerable the model is to this specific hallucination.
Our contributions are summarized as below:
\begin{itemize}
    \item We construct a fine-grained vision instruction dataset, RAI-30k. It contains multi-modal conversations focusing on specific vision relations in an image, enabling LVLMs to learn detailed vision features and spatial regions.
    \item We propose using the visual supervision to guide the model to focus on corresponding objects, thus generating more accurate responses and eliminating hallucination.
    \item We introduce a new hallucination benchmark, named RAH-Bench, which categorizes hallucinations into three distinct types and designs sub-metrics to enable more detailed analysis and assessment.
\end{itemize}

\begin{figure*}
  \centering
  \begin{subfigure}{0.85\linewidth}
    \includegraphics[width=\linewidth]{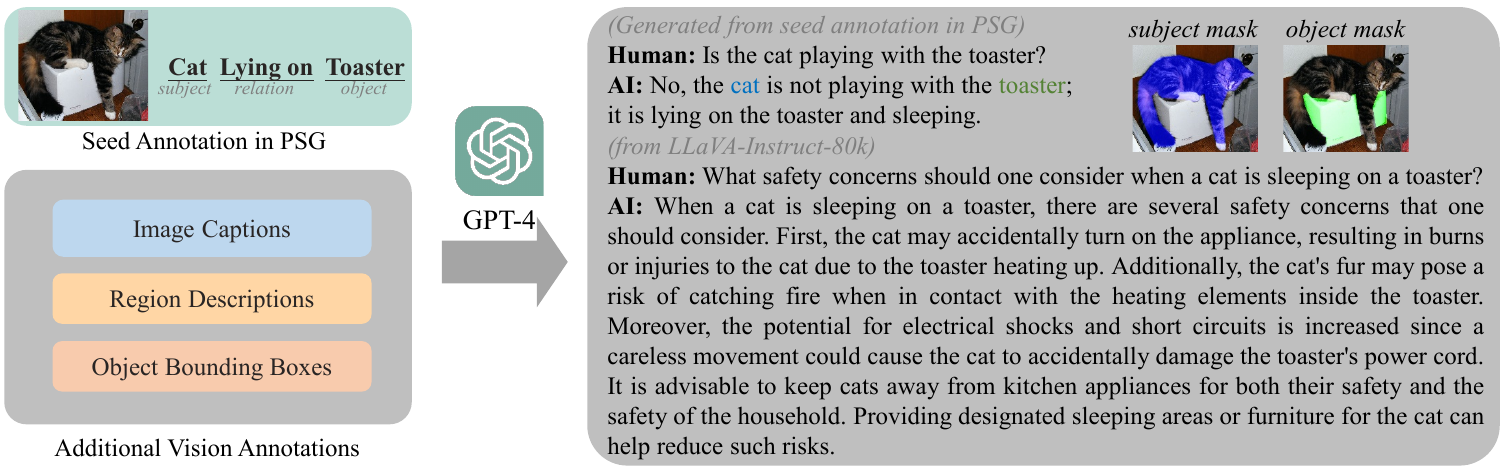}
    \caption{The illustration of the data construction pipeline with a seed annotation and GPT-4.}
    \label{fig:data_pipeline}
  \end{subfigure}
  \hfill
  \begin{subfigure}{\linewidth}
    \centering
    \includegraphics[width=0.8\linewidth]{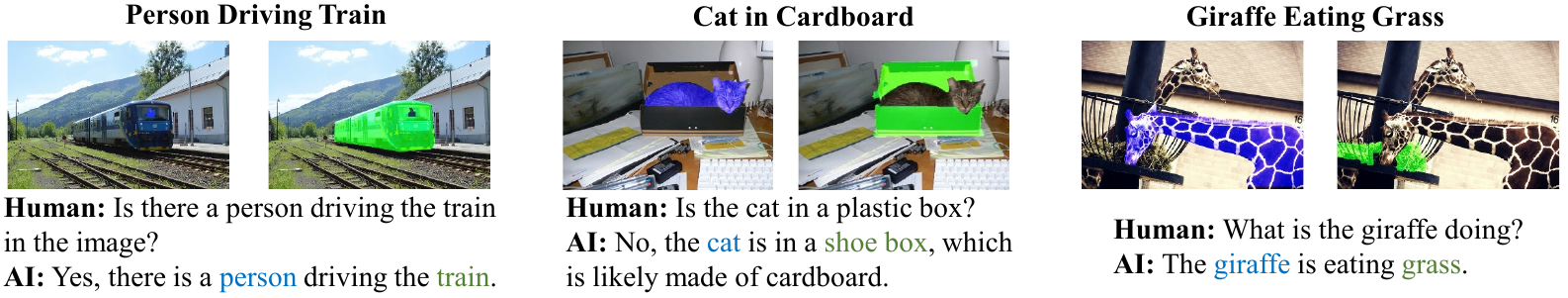}
    \caption{Examples of conversations generated from relation annotations.}
    \label{fig:data_sample}
  \end{subfigure}
  \caption{RAI-30k construction pipeline, and exampled of the generated conversations.}
  \vspace{-1em}
\end{figure*}
\section{Related Works}
\subsection{Large Vision-Language Models}
Inspired by the success of Large Language Models in NLP \cite{gpt4,llama,llama2,baichuan2}, researchers are now developing a wide range of Large Vision Language Models that have strong ability to perform a wide range of tasks with both image and text inputs. These models usually consists of a strong vision encoder \cite{clip} and a trained LLM \cite{llama2,vicuna}, and bridge them with a simple linear layer \cite{llava} or q-former \cite{blip2,instructblip,mplug}. 

In order to conduct various vision-language tasks, LVLMs need to be pretrained on large-scale image-text pairs, and then finetuned on vision instruction datasets.
Currently, most vision instruction datasets are constructed based on annotations in vision datasets \cite{coco,vg}. They utilize fixed templates \cite{instructblip,qwenvl} or GPT4 \cite{gpt4,llava,svit} to generate diverse data. However, most of these datasets are construct in a straight-forward manner, resulting in positive conversations and global descriptions. On the contrary, we construct our dataset with explicit focuses on the images, leading to fine-grained conversations and a high quality dataset.

\subsection{Combine LVLMs and Vision Models}
In order to accomplish more various tasks and provide detailed outputs, many recent works attempt to combine LVLMs with existing strong vision models. The most straight-forward solution is to let LVLMs generate commands to activate specific models \cite{hugginggpt,visualchatgpt,llavaplus}. These methods consider LVLMs and vision models as separate components, each with its inherent capabilities, and combine them with detailed rules.
Recently, LISA \cite{lisa} and ContextDET \cite{contextdet} propose to tune LVLMs and vision models together, so that the whole system has the capacity to flexibly predict detailed vision outputs based on the contexts. However, they mainly introduce new functions by appending additional models, while paying little attention on the abilities of LVLMs themselves.
On the contrary, our method mainly concentrates on how to use existing vision models and annotations to enhance LVLMs, so that they can generate accurate text as responses, mitigating vision hallucination.

\subsection{Hallucination in LVLMs}
LVLMs often prone to hallucination. Current works takes two kinds of approaches to mitigate it. One of them is to enrich the context with additional text inputs \cite{woodpecker}, akin to the use of external knowledge in LLMs \cite{knowledge1,knowledge2}. The other strategy focus on constructing high-quality instruction datasets with negative samples \cite{lrv_instruction} or iterative refinements \cite{vigc}. In addition to these approaches, we suggest that other than vision annotations, existing models and loss functions for visual tasks could also facilitate the training of LVLMs.

As to hallucination evaluation, current works usually provide overall metrics. POPE \cite{pope} performs binary classification tasks for easy assessment, while the questions in it share a consistent structural pattern. LRV-Instruction \cite{lrv_instruction} scores the responses with GPT-4. The question set is diverse, however, utilizing LLMs leads to high cost and unstable results. In this paper, we propose a benchmark that is both diverse and easy to evaluate. Moreover, we offer some more detailed sub-metrics as well, to uncover specific vulnerabilities in LVLMs.

\section{Method}
\label{sec:method}

To enhance the multi-modal context comprehension of LVLMs, we first construct a vision instruction dataset emphasizing detailed image analysis, and then utilize advanced vision models to facilitate LVLM training. It aims to improve LVLMs' ability to identify and interpret the key local regions in the image, thus generating more accurate responses. The following sections will elaborate on the development of RAI-30k dataset and the implementation of additional visual supervision separately.

\begin{figure*}
  \includegraphics[width=\linewidth]{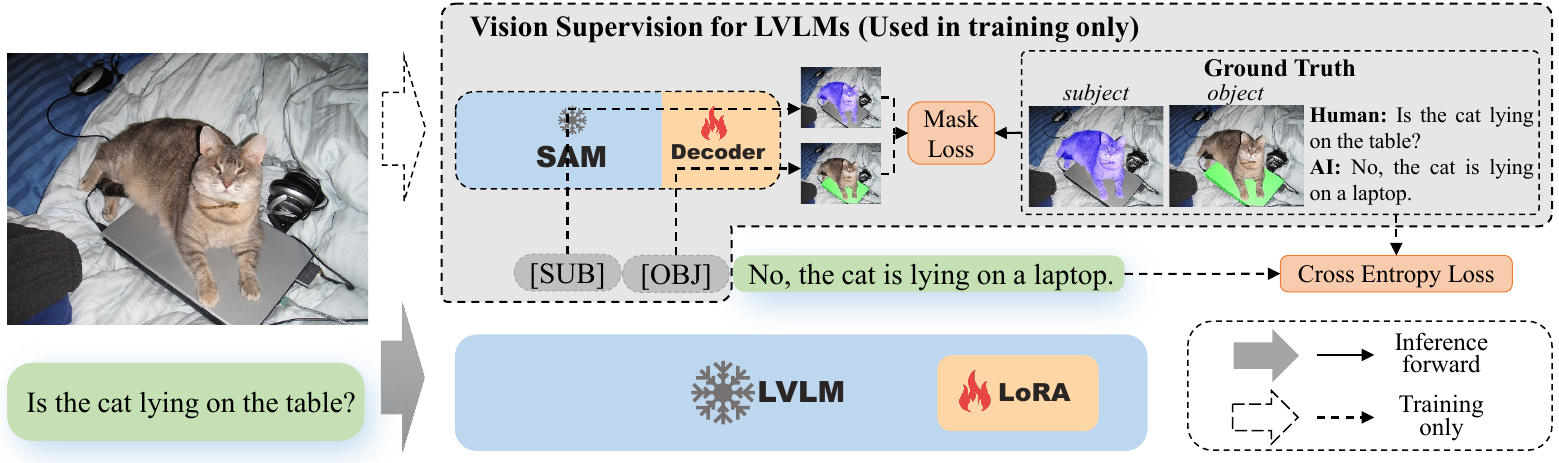}
  \caption{Illustration of how SAM facilitates vision instruction tuning.}
  \label{fig:loss}
\end{figure*}
\subsection{Relation-Associated Instruction Dataset}
\label{sec:data}
Conventional methods for constructing visual instruction datasets typically involve seeding all available captions and object annotations directly into GPT-4 \cite{gpt4,llava,svit}.
These approaches generally yield highly-consistent image-text pairs which are lack of distinctive focus and misleading context.
In contrast, daily discussions often center around specific elements, facts, or events, supplemented by related details. Therefore, we adopts a more nuanced strategy to generate conversations from seed topics. In this section, we take a vision relation annotation in panoptic scene graph dataset (PSG) \cite{psg} as the a seed topic.
The relation annotation comprises a subject, object, and their relation, providing detailed information to construct a single-round conversation that focuses exclusively on it.

As shown in Figure\ref{fig:data_pipeline}, we employ GPT-4 \cite{gpt4} to generate conversations around seed relation annotations, using a pool of three prompt variations to generate different questions, including yes/no interrogatives and open-ended questions starting with what, how, where, etc.
For each data sample, we randomly take one prompt template, fill the seed annotation in, and ask GPT-4 to generate a question-answer pair.
This approach ensures a wide range of conversational scenarios, mirroring the complexity and variability of real-world interactions with images. 
To further enrich the conversations with more details and ensure that they conform to the input image, we include detailed annotations from vision datasets \cite{coco,vg} in the prompts:
\begin{itemize}
    \item Captions of the entire image.
    \item Descriptions of specific regions, especially those overlapping with the subject or object in the seed annotation.
    \item A list of objects within the image, including their bounding boxes.
\end{itemize}

The resultant dataset, \emph{Relation-Associated Instruction-30k (RAI-30k)}, encompasses 29,712 data samples, each containing an image, a question-answer pair and the corresponding seed relationship annotation. The seed relationship includes the binary mask annotations of the subject and the object. Several generated examples are demonstrated in Figure~\ref{fig:data_sample}. To enrich our vision instruction data, we also append the multi-round conversations of the same images from LLaVA-Instruct 80k \cite{llava} to augment each data sample.
Overall, RAI-30k provides a diverse and fine-grained vision instruction dataset to train LVLMs for accurate, contextually relevant responses, effectively mitigating vision hallucination.

\begin{figure*}
  \centering
  \includegraphics[width=\linewidth]{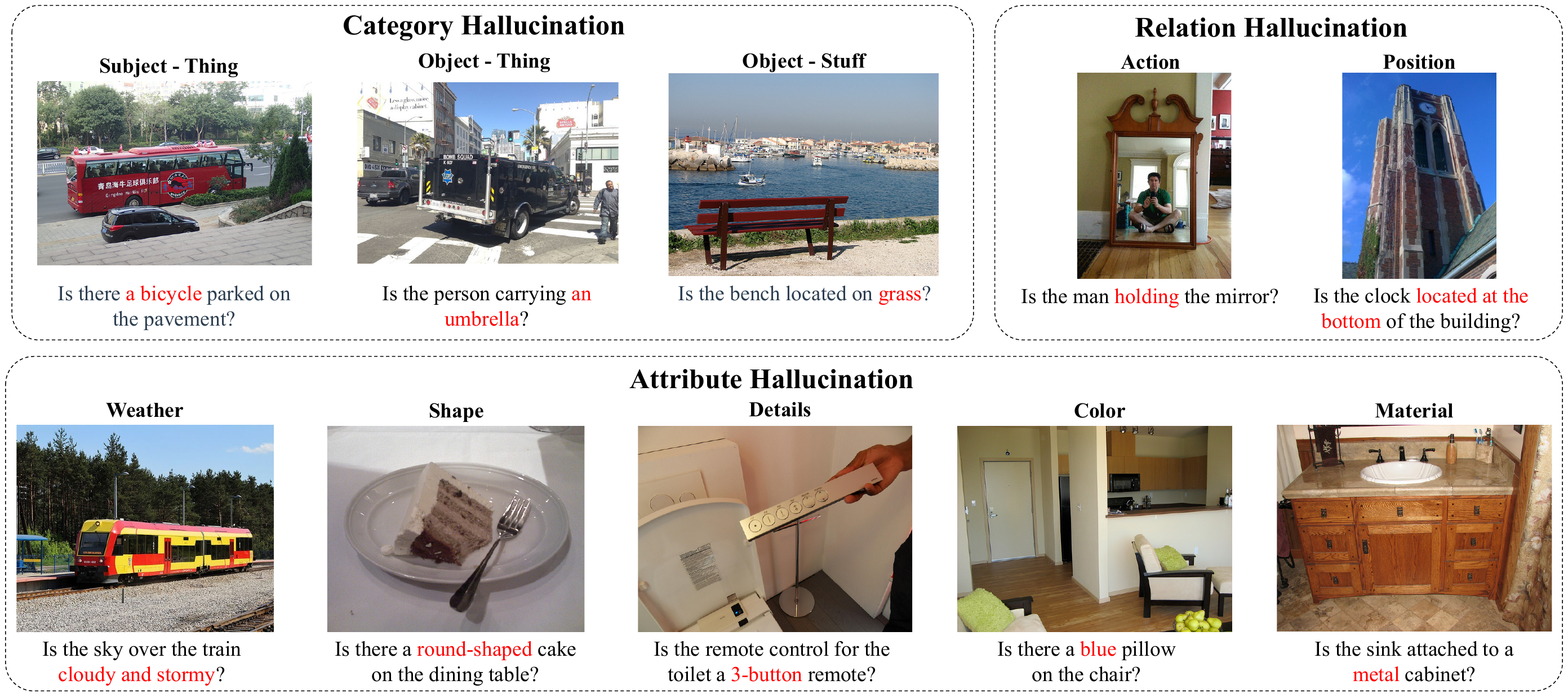}
  \caption{Examples of negative questions in RAH-Bench.}
  \label{fig:benchmark}
\end{figure*}

\subsection{Supervision with Visual Annotations}
\label{sec:loss}
In order to generate precise responses, human would first identify the key elements in context before answering. LVLMs are supposed to perform better in this way similarly. In RAI-30k, we have collected the critical relationship annotation as a supplement for each question, with detailed binary masks for the subject and the object.
These annotations can be utilized to explicitly guide LVLMs to attend to specific details in the image. Therefore, we integrate expert vision models and auxiliary mask prediction loss, including binary cross entropy and dice loss \cite{dice}, in the training stage of LVLMs, mitigating vision hallucination.

We exemplify our approach by incorporating SAM \cite{sam} in vision instruction tuning. SAM is a versatile segmentation model, designed to interpret various types of prompts and generate binary masks.
Take the training of LLaVA \cite{llava} as an example. We initialize LLaVA and SAM from their individually pretrained weights, and then attach SAM to facilitate instruction tuning of LLaVA. SAM takes the input prompts from LLaVA's output features, and computes mask prediction losses to guide LLaVA to attend to instances that are crucial to generate accurate responses.
More specifically, other than text responses, the LVLM generates two additional feature vectors $f_{sub}$ and $f_{obj}$ in its output sequence, referring to the two instances. These features are conditioned on the input image $x_{img}$ and question $x_{q}$. After processed through a linear layer $g(\cdot)$, these features are sent into SAM as the prompts to predict masks, and then be supervised with the ground truth masks in the seed relationship annotation. The whole process is shown in Figure~\ref{eq:mask}.
During training, we freeze most of the trained weight in LVLM and update LoRA \cite{lora} and the decoder in SAM for parameter-efficient tuning.
\begin{align}
  f_{sub}, f_{obj} &= \text{LLM}(x_{img}, x_{q})\\
  \text{Mask}_{sub} &= \text{SAM}(x_{img}, g(f_{sub}))\\
  \text{Mask}_{obj} &= \text{SAM}(x_{img}, g(f_{obj}))
  \label{eq:mask}
\end{align}

By appending two special tokens $[SUB]$ and $[OBJ]$ to extract corresponding features before the prediction,
the model is encouraged to identify the crucial instances based on only input context during training. Moreover, we modify the attention masks and position ids, so that these special tokens are processed alongside conventional tokens in a single forward pass. This modification does not have any influence on response generating in inference. 
After the training phase, SAM is discarded to ensure that the inference process remains unchanged. The LVLM generates text responses according to the input image and text tokens by conventional next token prediction.


\subsection{Relation-Associated Hallucination Benchmark}
\label{sec:benchmark}
In order to provide a detailed evaluation on vision hallucination, we introduce the \emph{Relation-Associated Hallucination Benchmark (RAH-Bench)}.
RAH-Bench contains 3,000 yes-or-no questions with their corresponding images. The images are from COCO validation set, while the questions are generated by GPT-4.
When evaluating, we simple parse the responses into binary classification results like in \cite{pope}, so that quantitative precision, recall and F1 score can be calculated.

Since LVLMs may generate responses that violate the image contents from different perspectives, we further divide the negative questions into three subsets to better reveal which questions are more likely to cause vision hallucination. Each subset contains 500 questions with misleading statements in different aspects, including:
\begin{itemize}
    \item \textbf{Categorical Hallucination}: LVLMs identify nonexistent object categories or incorrect background categories in the given image.
    \item \textbf{Attribute Hallucination}: The object categories identified by LVLMs are accurate, while the descriptions of these objects' attributes (such as color, shape, material, content, etc.) are wrong.
    \item \textbf{Relation Hallucination}: All objects and their attributes are described correctly, but the relationships among them (such as human-object interactions or relative positions) do not align with the actual image content.
\end{itemize}

Figure~\ref{fig:benchmark} presents some examples of different types. To reveal the susceptibility of LVLMs to different types of misleading queries, we introduce False Positive Rates (FP) for each subset, which represents the probability that this hallucination type would occur.
\begin{equation}
  \text{FP} = \frac{\text{Num. of False Positives}}{\text{Num. of Samples}}.
\end{equation}

Compared to previous benchmark in vision hallucination \cite{pope}, RAH-Bench has the following advantages. Firstly, RAH-Bench feeds GPT-4 with more visual annotations, leading to complex and diverse questions. Secondly, RAH-Bench blends truths and misleads in a single question, posing a greater challenge for distinction. Finally, the specific hallucination types and the additional sub-metric can provide a deeper evaluation on how LVLMs are vulnerable to different vision hallucination.


\begin{table*}
  \centering
  \begin{tabular}{@{}c@{}c|ccc|ccc|ccc|c@{\hspace{0.8\tabcolsep}}c@{\hspace{0.8\tabcolsep}}c@{\hspace{0.8\tabcolsep}}c@{}}
    \hline
    \multicolumn{2}{c|}{Model} &
    \multicolumn{3}{c|}{POPE Random} &
    \multicolumn{3}{c|}{POPE Popular} &
    \multicolumn{3}{c|}{POPE Adversarial} &
    \multicolumn{4}{c}{RAH-Bench} \\
    \hline
    Name & Base LLM &
    P$^\uparrow$ & R$^\uparrow$ & F1$^\uparrow$ &
    P$^\uparrow$ & R$^\uparrow$ & F1$^\uparrow$ &
    P$^\uparrow$ & R$^\uparrow$ & F1$^\uparrow$ &
    FP$^{\downarrow}_{cat}$ & FP$^{\downarrow}_{attr}$ & FP$^{\downarrow}_{rela}$ & F1$^\uparrow$\\
    \hline
    mPLUG-owl & vicuna-7B &
    54.9 & 99.3 & 70.7 &
    50.4 & 99.3 & 66.8 &
    50.3 & 99.4 & 66.8 &
    81.2 & 93.6 & 86.4 & 69.3 \\
    +ours &  & 
    58.5 & 99.2 & \textbf{73.6} &
    55.5 & 99.2 & \textbf{69.0} &
    52.1 & 99.3 & \textbf{68.3} &
    \textbf{70.6} & \textbf{80.6} & \textbf{76.2} & \textbf{71.6} \\
    \hline
    InstructBLIP & vicuna-7B   & 
    93.2 & 87.5 & \textbf{90.2} &
    82.2 & 87.5 & \textbf{84.8} &
    76.7 & 87.5 & 81.7 &
    9.2 & 12.0 & 32.2 & 89.1 \\
    +ours &  &
    95.6 & 84.9 & 90.0 &
    84.4 & 84.9 & 84.7 &
    80.3 & 84.9 & \textbf{82.5} &
    \textbf{5.4} & \textbf{5.8} & \textbf{23.4} & \textbf{89.6}\\
    InstructBLIP & vicuna-13B   & 
    83.1 & 94.1 & 88.2 &
    74.8 & 94.1 & 83.4 &
    66.9 & 94.1 & 78.2 &
    20.2 & 35.8 & 46.8 & 84.7 \\
    +ours &  &
    96.1 & 84.2 & \textbf{89.8} &
    85.5 & 84.2 & \textbf{84.8} &
    81.5 & 84.2 & \textbf{82.8} &
    \textbf{5.2} & \textbf{4.8} & \textbf{22.4} & \textbf{88.7}\\
    \hline
    LLaVA & LLaMA2-7B &
    55.7 & 99.7 & 71.5 &
    51.4 & 99.6 & 67.8 &
    50.7 & 99.7 & 67.3 &
    66.2 & 65.6 & 61.8 & \textbf{73.3} \\
    +ours &  & 
    69.0 & 95.3 & \textbf{80.1} &
    60.6 & 95.3 & \textbf{74.1} &
    57.0 & 95.4 & \textbf{71.4} &
    \textbf{31.4} & \textbf{28.0} & \textbf{24.4} & 71.4 \\
    LLaVA & LLaMA2-13B &
    60.1 & 98.1 & 74.6 &
    59.0 & 98.4 & 73.8 &
    55.3 & 98.2 & 70.7 &
    69.0 & 83.6 & 77.6 & 71.8 \\
    +ours &  & 
    84.2 & 85.3 & \textbf{84.7} &
    76.7 & 86.3 & \textbf{81.2} &
    67.4 & 85.2 & \textbf{75.3} &
    \textbf{24.0} & \textbf{21.6} & \textbf{28.4} & \textbf{80.2}\\
    \hline
  \end{tabular}
  \caption{Experiments on different up-to-date LVLMs.}
  \label{tab:main}
\end{table*}
\section{Experiments}
\subsection{Main Results}
In this paper, we provide a data construction pipeline and a training algorithm to mitigate vision hallucination. These methods are model-independent, thus can be applied to most up-to-date LVLMs \cite{llava,mplug,instructblip}. We conduct experiments with several different models to validate our efficacy. These LVLMs are different in base LLMs \cite{llama2,vicuna}, model sizes, and detailed structures of the adapter between the vision encoder and the LLM.

As shown in Table~\ref{tab:main}, most existing LVLMs have high recall and relatively low precision, indicating the existence of vision hallucination. With FP metrics in the subsets of RAH-Bench, we can see that LLaVA and mPLUG-owl provide positive responses to more than half of the misleading questions. Moreover, 13B models do not necessarily perform better than 7B models. With our method equipped, the F1 score of LLaVA-13B is improved by +8.4\% on RAH-Bench, and by +10.1\%, +7.4\%, and +3.9\% with different evaluation settings on POPE \cite{pope}, indicating a geneal improvement across different metrics. As to mPLUG-owl, the F1 score is increased by +2.3\%. This smaller improvement may be due to the fact that we tune all these models with the same training settings and hyper-parameters as in LLaVA, which may not be optimal for other LVLMs. Nevertheless, these experiments validate that our method is versatile to mitigate vision hallucination in different LVLMs.

InstructBLIP \cite{instructblip} achieves significantly better performance than other models. This is mainly due to its comprehensive instruction dataset constructed from 20+ vision datasets with various annotations. This dataset already contains some detailed information. Even so, when testing on RAH-Bench, we find that InstructBLIP is more vulnerable to relation hallucination, resulting in a FP significantly higher than other types. Queries within this subset invariably combine accurate instance descriptions with incorrect relational details between them, therefore requiring reasoning at a higher semantic level. Therefore, we may conclude that InstructBLIP is better at identifying facts than complex reasoning. This can be enhanced with our method. Since InstructBLIP only supports single-round conversation, we only tune them on the generated question-answer pairs. As shown in Table~\ref{tab:main}, our tuning preserves the performance on the random and popular split within POPE, and gains a significant improve for the harder adversarial split (+0.8\%/+4.6\% for 7B/13B) and our RAH-Bench (+0.5\%/+4.0\%), which means our tuning improves the reasoning ability in InstructBILP and enables it to discriminate those easy-to-confuse statements.

\begin{table*}
  \centering
  \begin{tabular}{@{}l@{\hspace{0.75\tabcolsep}}l|ccc|ccc|ccc|c@{\hspace{0.65\tabcolsep}}c@{\hspace{0.65\tabcolsep}}c@{\hspace{0.65\tabcolsep}}c@{}}
    \hline
    &Experiment &
    \multicolumn{3}{c|}{POPE Random} &
    \multicolumn{3}{c|}{POPE Popular} &
    \multicolumn{3}{c|}{POPE Adv.} &
    \multicolumn{4}{c}{RAH-Bench} \\
    &&
    P$^\uparrow$ & R$^\uparrow$ & F1$^\uparrow$ &
    P$^\uparrow$ & R$^\uparrow$ & F1$^\uparrow$ &
    P$^\uparrow$ & R$^\uparrow$ & F1$^\uparrow$ &
    FP$^{\downarrow}_{cat}$ & FP$^{\downarrow}_{attr}$ & FP$^{\downarrow}_{rela}$ & F1$^\uparrow$\\
    \hline
    1&Baseline &
    60.1 & 98.1 & 74.6 &
    59.0 & 98.4 & 73.8 &
    55.3 & 98.2 & 70.7 &
    69.0 & 83.6 & 77.6 & 71.8 \\
    \hline
    2&Ours & 
    77.8 & 90.5 & \textbf{83.7} &
    72.2 & 89.7 & \textbf{80.0} &
    64.6 & 90.5 & \textbf{75.4} &
    \textbf{32.8} & \textbf{34.6} & \textbf{38.6} & 78.7 \\
    3&w/o Detailed Annotations  &
    60.8 & 98.7 & 75.3 &
    58.6 & 98.7 & 73.5 &
    55.3 & 98.5 & 70.8 & 
    73.6 & 84.8 & 82.0 & 70.9 \\
    4&w/o Varied prompts & 
    75.9 & 91.6 & 83.0 &
    70.2 & 92.2 & 79.7 &
    63.4 & 92.3 & 75.2 &
    35.2 & 37.2 & 41.4 & \textbf{79.5} \\
    5&w/o Data in LLaVA & 
    72.9 & 95.0 & 82.5 &
    68.0 & 94.9 & 79.2 &
    61.8 & 94.7 & 74.8 &
    44.8 & 46.4 & 55.6 & 72.5 \\
    \hline
  \end{tabular}
  \caption{Experiments on the construction process of RAI-30k.}
  \label{tab:data}
\end{table*}

\begin{table*}
  \centering
  \begin{tabular}{@{}cc|ccc|ccc|ccc|c@{\hspace{0.8\tabcolsep}}c@{\hspace{0.8\tabcolsep}}c@{\hspace{0.8\tabcolsep}}c@{}}
    \hline
    Vision Model & Epochs & 
    \multicolumn{3}{c|}{POPE Random} &
    \multicolumn{3}{c|}{POPE Popular} &
    \multicolumn{3}{c|}{POPE Adversarial} &
    \multicolumn{4}{c}{RAH-Bench} \\
     & &
    P$^\uparrow$ & R$^\uparrow$ & F1$^\uparrow$ &
    P$^\uparrow$ & R$^\uparrow$ & F1$^\uparrow$ &
    P$^\uparrow$ & R$^\uparrow$ & F1$^\uparrow$ &
    FP$^{\downarrow}_{cat}$ & FP$^{\downarrow}_{attr}$ & FP$^{\downarrow}_{rela}$ & F1$^\uparrow$\\
    \hline
     &  1 & 
    77.8 & 90.5 & 83.7 &
    72.2 & 89.7 & 80.0 &
    64.6 & 90.5 & 75.4 &
    32.8 & 34.6 & 38.6 & 78.7 \\
    \checkmark & 1  &
    84.2 & 85.3 & 84.7 &
    76.7 & 86.3 & 81.2 &
    67.4 & 85.2 & 75.3 &
    24.0 & 21.6 & 28.4 & 80.2\\
    \checkmark & 3  &
    87.2 & 88.2 & 87.7 &
    76.2 & 88.7 & 81.9 &
    67.7 & 88.3 & 76.6 &
    15.6 & 10.2 & 16.6 & 79.5 \\
    \checkmark & 5  &
    87.4 & 88.7 & 88.0 &
    77.0 & 88.9 & 82.5 &
    69.5 & 88.7 & 77.9 &
    12.6 & 5.8 & 15.6 & 80.0\\
    \hline
  \end{tabular}
  \caption{Experiments on mask prediction losses as auxiliary supervision.}
  \label{tab:loss}
\end{table*}
\subsection{Ablation on Data Construction}
In Section~\ref{sec:data}, we have introduced the data construction pipeline for RAI-30k. The whole pipeline can be simply summarized into two steps: generating relation-associated question-answer pairs, and append existing multi-round conversations in LLaVA-Instruct-80k.

First we ablate on how to generate question-answer pairs. Providing more abundant annotations is important, since the seed relationships in PSG \cite{psg} only contain simple category names like \emph{people drive car}. Data generated with these only would be less relevant to detailed image contents. As shown in Table~\ref{tab:data} L3, tuning with these uninformative data leads to degradation in model performance. Meanwhile, captions from COCO \cite{coco} and region descriptions from Visual Genome \cite{vg} are more detailed and informative. With them as additional context, GPT4 can pose questions that target specific fact in the given image. When testing with RAH-Bench, it leads to +6.9\% improvement on F1 score, and False Positive Rates generally decreases for all hallucination types.

In order to generate more diverse questions, we design three different prompts for GPT-4, corresponding to different question types: questions that should be answered with yes / no, and wh-questions formulated to elicit detailed information. Though the evaluation metric only requires binary classification results, by comparing L2 and L4 in Table~\ref{tab:data}, manually prompting GPT-4 to generate more diverse questions results in slightly higher F1 scores on all splits in POPE.
We suggest that designing more question types (e.g. providing more prompts) will further improve LVLMs' performance, as the training data becomes more diverse.

Finally, we investigate the impact of incorporating multi-round data from LLaVA-Instruct-80k \cite{llava} into our generated dataset. Our question-answer pairs predominantly concentrate on specific relationships, contrasting with the broader range of instances and common knowledge presented in LLaVA. As shown in L2 and L5 in Table~\ref{tab:data}, augmenting RAI-30k with more diverse conversational leads to a large improvement in performance on RAH-Bench (+6.2\%), a benchmark with diverse questions, while a modest improvement on the more uniform POPE benchmark.

\subsection{Ablation on Vision Supervision}
In this section, we provide additional experiments to dive deeper into the function of SAM and the auxiliary mask prediction loss. As shown in Table~\ref{tab:loss}, adding these vision components in training leads to 1.5\% higher F1 score on RAH-Bench than tuning with RAI-30k with a conventional manner, indicating that the expert vision model helps alleviate vision hallucination. Continue training with more epochs would gain further improvements. However, excessive training epochs on this dataset could result in overfitting to the binary classification task, consequently impairing the general ability of LVLMs. We evaluate this ability with LLaVA-Bench (In-the-Wild) as in Table~\ref{tab:llava}, which is a benchmark to assess models' robustness to different prompts. Training for one epoch does not affect the general capacities of LLaVA much, therefore we adopt this duration as the default training configuration.

\begin{table}
  \centering
  \begin{tabular}{@{}l|c@{\hspace{0.8\tabcolsep}}c@{\hspace{0.8\tabcolsep}}c|c@{}}
    \hline
    Model & Conv.$^\uparrow$ & Detail$^\uparrow$ & Reasoning$^\uparrow$ & Overall$^\uparrow$\\
    \hline
    LLaVA-13B        & 50.6 & 59.0 & 80.1 & 66.1 \\
    + ours (1 epoch) & 46.9 & 64.4 & 80.9 & 66.5 \\
    + ours (5 epoch) & 46.9 & 57.8 & 77.3 & 63.3 \\
    \hline
  \end{tabular}
  \caption{Assessment of models' robustness on LLaVA-Bench (In-the-Wild), including the ability on conversation, detailed description and complex reasoning.}
  \label{tab:llava}
\end{table}


Moreover, by comparing the detailed metrics, we further find that after tuning, our model obtains a significant enhancement in its ability to generate detailed description, alongside a modest improvement in complex reasoning. This may be attributed to our fine-grained vision instruction dataset and mask prediction supervision, which provides additional evidences that the model trained with our method can better perceive detailed image contents and perform multi-modal deduction.


\subsection{Visualization}
\label{sec:vis}
\begin{figure*}
  \centering
  \includegraphics[width=\linewidth]{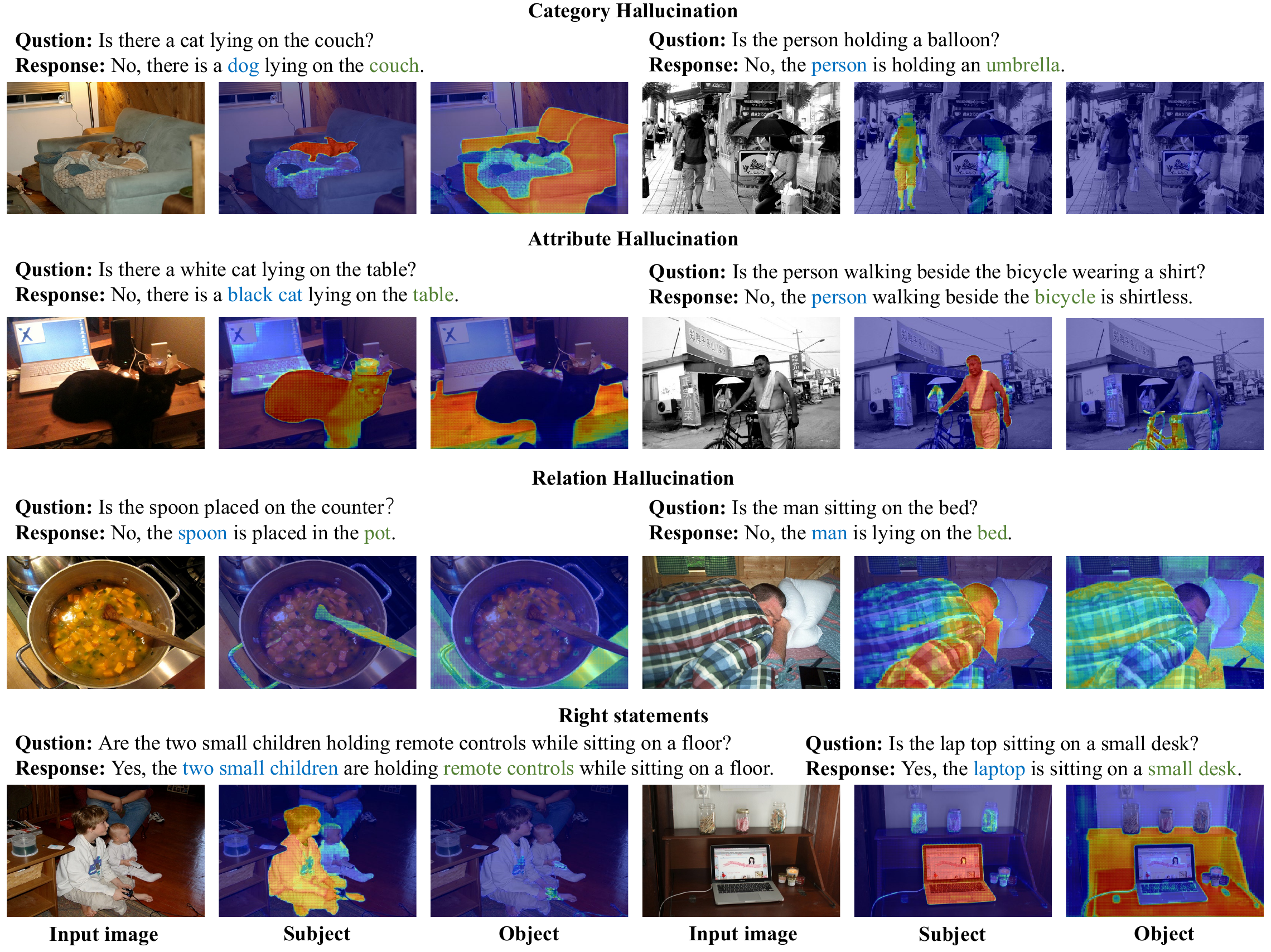}
  \caption{Illustration of how our model helps LVLMs identify the key items and answer the question. In addition to the model's responses, we draw the mask that the model would attend to.}
  \label{vis:mask}
\end{figure*}

To better demonstrate how our method helps LVLMs find crucial instances, we provide the following analysis.

Since SAM is attached and LVLMs are forced to predict the right masks in the seed relation annotation, we may draw the masks out to see if the LVLMs do recognize the crucial items. As shown in Figure~\ref{vis:mask}, LVLMs are able to attend to the key items associated with the question. If the question provides correct descriptions for the subject and object (e.g. right questions or relation hallucination), the model can identify them. If the question provides misleading questions about them (e.g. object hallucination and attribute hallucination), the model will attend to the content that can point out the contradiction, or just predict low scores on the whole output mask. The mask supervision loss encourages the LVLMs to generate more precise responses, and also makes their behavior more explainable.
Note that we only visualize these images to visualize the model behavior. In practical inference, LVLMs just conduct conventional next-token generation.
Overall, these illustrations demonstrate that our model can associate the provided text with image contents, and infer final responses according to them.


\section{Conclusion}
In this paper, we construct a fine-grained vision instruction dataset RAI-30k that focuses on specific details in images, and propose a methodology to enhance LVLMs with supervision from delicated losses in expert vision models. These could help mitigate vision hallucination in different LVLMs. Moreover, in order to provide a more detailed evaluation on vision hallucination, we develop RAH-Bench. This benchmark includes various categorized subsets specifically designed to assess the severity of different types of hallucination in LVLMs. We hope our work provides a solid foundation for further research in LVLM training and vision hallucination.

While we have validated the efficacy of our method, it is important to acknowledge certain limitations in our research. Firstly, the dataset we constructed is not that large. Expanding it from more seed annotations could potentially enhance performance.
Secondly, though we design a general method to facilitate vision instruction tuning with expert vision models,
our training approach is somewhat coupled with the data format in RAI-30k, such as \emph{subject} and \emph{object} in relation annotations. In the future, we will evolve our method to accommodate a broader range of vision models and more versatile annotation formats.
{
    \small
    \bibliographystyle{ieeenat_fullname}
    \bibliography{main}
}


\end{document}